\documentclass[letterpaper, 10 pt, journal, twoside]{IEEEtran}
\IEEEoverridecommandlockouts                              % This command is only needed if
\usepackage{amsfonts}
\usepackage{amsmath}
\usepackage{graphicx}
\usepackage{subcaption}
\usepackage[skip=0pt,labelfont=bf]{caption}
\usepackage{wrapfig}
\usepackage{algorithm}
\usepackage{algorithmic}
\usepackage{textcomp}
\newcommand\norm[1]{\left\lVert#1\right\rVert}
\usepackage{cite} % Group multiple citations into brackets

\DeclareMathOperator*{\argmin}{arg\,min}
\usepackage{url}

\title{\LARGE \bf
Building 3D Object Models during Manipulation\\ by Reconstruction-Aware Trajectory Optimization
}

\author{Kanrun Huang$^{1}$ and Tucker Hermans$^{1}$% <-this % stops a space
\thanks{$^{1}$Kanrun Huang and Tucker Hermans are with the School of Computing and the Robotics Center, University of Utah, UT 84112, USA.
        {\tt\small email: u1063462@utah.edu,thermans@cs.utah.edu}}%
}

\begin{document}
\maketitle
\thispagestyle{empty}
\pagestyle{empty}

%%%%%%%%%%%%%%%%%%%%%%%%%%%%%%%%%%%%%%%%%%%%%%%%%%%%%%%%%%%%%%%%%%%%%%%%%%%%%%%%
\begin{abstract}
  Object shape provides important information for robotic manipulation; for instance, selecting an effective grasp depends on both the global and local shape of  the object of interest, while reaching into clutter requires accurate surface geometry to avoid unintended contact with the environment. Model-based 3D object manipulation is a widely studied problem; however, obtaining the accurate 3D object models for multiple objects often requires tedious work. In this letter, we exploit Gaussian process implicit surfaces (GPIS) extracted from RGB-D sensor data to grasp an unknown object. We propose a reconstruction-aware trajectory optimization that makes use of the extracted GPIS model plan a motion to improve the ability to estimate the object's 3D geometry, while performing a pick-and-place action. We present a probabilistic approach for a robot to autonomously learn and track the object, while achieve the manipulation task.
  We use a sampling-based trajectory generation method to explore the unseen parts of the object using the estimated conditional entropy of the GPIS model.
We validate our method with physical robot experiments across eleven different objects of varying shape from the YCB object dataset. Our experiments show that our reconstruction-aware trajectory optimization provides higher-quality 3D object reconstruction when compared with directly solving the manipulation task or using a heuristic to view unseen portions of the object.
\end{abstract}
%%%%%%%%%%%%%%%%%%%%%%%%%%%%%%%%%%%%%%%%%%%%%%%%%%%%%%%%%%%%%%%%%%%%%%%%%%%%%%%%
%%%%%%%%%%%%%%%%%%%%%%%%%%%%%%%%%%%%%%%%%%%%%%%%%%%%%%%%%%%%%%%%%%%%%%%%%%%%%%%%
\vspace{-16pt}
\section{Introduction}
\vspace{-4pt}
Object shape provides highly informative information for autonomous, robot manipulation tasks like grasping and pushing or robot vision tasks like object detection and tracking. However, providing three-dimensional (3D) models of all objects of interest to a robot may require significant amounts of time or even be impossible in scenarios such as search and rescue in disaster areas where rubble and other debris can not be known exactly a priori. As such, researchers have proposed an abundance of approaches to perform autonomous shape estimation using RGB-D sensors such as the Microsoft Kinect, tactile sensors, or a combination of the two~\cite{sundaralingam-auro2018-in-grasp-optimization,c26,c27,c28}.

Manipulation planning and shape estimation are fundamentally related--more accurate object shape helps to improve planning~\cite{c33}, while manipulation of an object provides additional sensor measurements~\cite{c25} improving object model quality.

\begin{figure}[thpb]
	\centering
	\includegraphics[width=\linewidth]{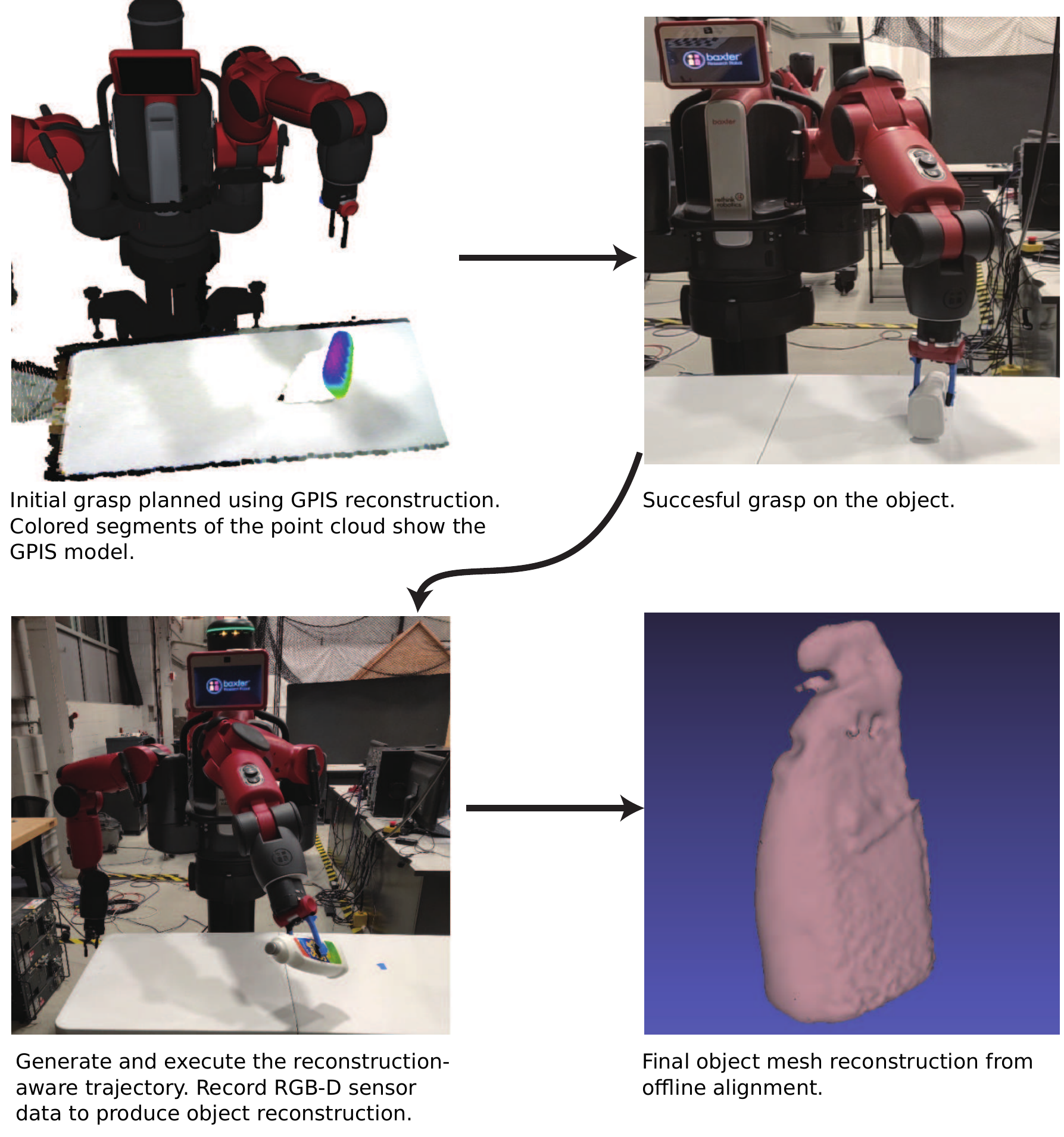}
	\caption{Overview of the proposed approach to autonomous object exploration and reconstruction.}
	\label{fig:grasp-overview}
\end{figure}
In this letter, we investigate the problem of constructing a 3D model of unknown object jointly with planning and executing a pick-and-place trajectory for the object. Figure~\ref{fig:grasp-overview} shows an overview of our approach. We first segment the object from a single view using the point cloud from an RGB-D sensor and estimate a probabilistic model of the object represented as a Gaussian process implicit surface (GPIS). We then use this probabilistic model to plan a grasp and subsequent robot-and-object trajectory to a predefined placement location. We use a probabilistic planner for the trajectory optimization, where we explicitly reason about the viewpoint of the object in the camera frame. Using the uncertainty associated with the GPIS object model we can compute the conditional entropy of the posterior to plan motions that explore regions of the object with high uncertainty. We then execute the pick-and-place trajectory and use the data collected during execution to reconstruct the object model.

This letter propose the first motion planner that takes into account 3D reconstruction of an unknown object, while manipulating the object in a goal-directed manner. We validate this contribution by comparing to an uninformed planner and heuristic planning approach with physical experiments on a physical Baxter robot with a set of ten objects from the YCB object set~\cite{c11}. Our results show that jointly considering manipulation and reconstruction provide higher-quality object models than the comparison approaches.

We organize the rest of the letter as follows. We discuss related work on autonomous 3D reconstruction and GPIS-based grasp planning in Section~\ref{sec:related-work}.
We review the GPIS shape representation in Section~\ref{sec:gpis}. In Section~\ref{sec:planning} we provide the details for our reconstruction-aware motion planning algorithm. We explain our experimental design and results, as well as implementation details in Section~\ref{sec:experiments}. We conclude and provide a brief discussion of future research directions in Section~\ref{sec:conclusion}.

%%%%%%%%%%%%%%%%%%%%%%%%%%%%%%%%%%%%%%%%%%%%%%%%%%%%%%%%%%%%%%%%%%%%%%%%%%%%%%%%

\section{Related Work}\label{sec:related-work}%\vspace{-12pt}
Reconstruction of 3D objects and indoor environments has been widely studied in robotics~\cite{c35,c36,c37,c38}.
Several planning approaches have been proposed to improve reconstruction performance in 3D simultaneous localization and mapping tasks.
These approaches base planning on improving some measure of information to guide a mobile robot to explore areas with high uncertainty~\cite{c13,c14}. Julian et al.~\cite{c14} assume known robot poses, and rely on an occupancy grid map representation. The algorithm integrates over an information gain function with an inverse sensor model at its core. In a similar vain the work of~\cite{c13} uses the gradient of the Cauchy-Schwarz quadratic mutual information to perform trajectory planning, as it allows more efficient computation than the standard mutual information measure.

Typical approaches to object reconstruction use variants of iterative closest point (ICP) to perform front-end alignment for reconstruction. Xie et al.~\cite{c17} use ICP to track object point cloud to build 3D models and use RANSAC for initialization of ICP, and they test the result on the open RGB-D dataset from~\cite{c32}, where the mask of the object is well defined.
Newcombe et al.~\cite{c18} demonstrated their famous KinectFusion approach that dense reconstruction is possible in real-time by using a Microsoft Kinect sensor. To represent the geometry, Newcombe et al. employ a truncated signed distance function (TSDF) and use ICP in a coarse-to-fine manner to estimate the camera pose in real-time. Krainin et al.~\cite{c15} develop an approach to build 3D models of unknown objects based on a depth camera observing the robot’s hand after moving an object to a certain pose. The approach integrates both shape and appearance information into an articulated ICP approach to track the robot$^{\prime}$s manipulator and the object while improving 3D model of an object. This work was extended to plan successive grasps in lifting the object that consider the occlusion caused by the robot's hand into account~\cite{krainin-icra2011-nbv}; however, the specific trajectories used in generating object motion were heuristically generated without a specific manipulation goal.

Gaussian process object models have been used as an important tool to predict the uncertainty of object shape for several different robotic tasks~\cite{c8,c9,c20}. These were first used in planning grasps~\cite{c8,c9}. However, more recent works have used tactile information to improve estimates of object geometry with the use of GPIS models. Several examples along this line of work include~\cite{c20,c1,c2,c4}. Yi et al.~\cite{c20} use a 2.5D GP model to iteratively select discrete locations to actively guided touches of the unknown object's part. Other approaches try multiple grasps to obtain tactile points, for instance~\cite{c4,c1,c2} use both tactile and RGB-D sensors to build 3D object models; however, these methods may require many (e.g. more than 50 in~\cite{c2}) touches to fully cover an object.

Interactive perception~\cite{bohg-tro2017-interactive-perception} has shown success in many different applications, most significantly object segmentation~\cite{c23,c34}. Most similar to our planning approach Van Hoof et al.~\cite{c34} plan a pushing action from a discrete set using an information-gain metric over their probabilistic segmentation. Ma et.al~\cite{c5} also perform active manipulation enabling reconstruction from tactile and 3D geometric information during manipulation.

%%%%%%%%%%%%%%%%%%%%%%%%%%%%%%%%%%%%%%%%%%%%%%%%%%%%%%%%%%%%%%%%%%%%%%%%%%%%%%%%

\section{Gaussian Process Implicit Surfaces\\ for Shape Representation}\label{sec:gpis}
We briefly review using Gaussian process regression to probabilisticly model implicit surface functions~\cite{c16,c8,c9,c6}. We can then use this probabilistic implicit surface model to estimate whether a specific point of interest in three dimensional space lies on the object surface.

We formally define our probabilistic implicit surface by
\begin{align}
f(x)\left\{
\begin{array}{rcl}
>\eta      &      & \mbox{if x is outside the object}\\
\le|\eta|&& \mbox{if x is on the surface of the object}\label{eq:implicit-surf}\\
<-\eta     &      & \mbox{if x is inside the object}
\end{array} \right.
\end{align}
where we set a small confidence threshold $\eta$ to decide whether an estimated point lies on the surface or not. If we shrink $\eta$ to zero, then Eq.~\ref{eq:implicit-surf} collapses to an equality comparison and we recover the classic (deterministic) implicit surface model.

We can represent the function \(f(x)\) using a Gaussian process by collecting a training set $S_{train}$ consisting of a set of locations in 3D space $X=\{x_0,x_1......x_n\}$, $x_i\in\mathbb{R}^3$ of points lying on the surface of the object, which we provide associated labels $Y=\{y_0,y_1......y_n\}$, $y_i = 0$.

A GP is composed of mean function, $m(\cdot)$, kernel function, $k(\cdot,\cdot)$, and output Gaussian noise $\sigma_n$,
\begin{align}
f(x) \sim GP(m(\cdot,\cdot),k(\cdot,\cdot)) \\
y = f(x) + \sigma_n
\end{align}
The mean and covariance of a query point $x^*_q\in\mathbb{R}^3$ in an estimated set $S_{estimated}=\{x^*_0,x^*_1......x^*_n\}$, are estimated (following~\cite{c21}) as
\begin{align}
\mu(x^*_q) = m(x^*_q) + k^{*T}K^{-1}(y_D-m_{f,D}) \\
\sigma^2(x^*_q)  = k^{**} - k^{*T}K^{-1}k^* \\
K \in \mathbb{R}^{N\times N}  \qquad K_{ij} = k(x_i,x_j)\\
k^* \in \mathbb{R}^{N\times 1}  \qquad k^* = k(X,x^*_q)\\
k^{**} \in \mathbb{R} \qquad k^{**} = k(x^*_q,x^*_q)
\end{align}
In our case, we use the squared-exponential kernel for Gaussian process; defined as
\begin{equation}
    k(x_i,x_j) = \sigma^{2}(-\frac{1}{2L^2}\norm{x_i-x_j}^2)
\end{equation}
where $\sigma$ is the intensity and $L$ is the length scale.

We found that using the zero mean function worked best for our purposes. Following the approach in~\cite{c16} we examined the use of an elliptical function as mean to better exploit common object symmetries, but found these to overfit to the prior.

 In practice, we estimate the surface model from a single RGB-D frame. We assume that there is only one object lying on the table and it is reachable for robot. We use the random sample consensus (RANSAC) to fit and segment the table plane in the point cloud. Then we build a KD-tree for the remaining points in the point cloud at each time frame and perform Euclidean cluster to obtain points relating to the object as the training set $S_{train}$ for the Gaussian process. We execute all of these steps using tools from the Point Cloud library (PCL)~\cite{c31}.

%%%%%%%%%%%%%%%%%%%%%%%%%%%%%%%%%%%%%%%%%%%%%%%%%%%%%%%%%%%%%%%%%%%%%%%%%%%%%%%%

\section{Manipulation Planning and Reconstruction}\label{sec:planning}
We decompose the discussion of our 3D model reconstruction process broadly into two tasks: (1) grasping the object using a parallel-jaw gripper based on the extracted GPIS representation and (2) planning and executing a reconstruction-aware trajectory to the object's goal pose, while capturing the associated sensor data. We delve into the details of each component in the remainder of this section.
\vspace{-10pt}
\subsection{GPIS-Based Grasp Trajectory Planning}\label{s41}
Prior to grasping the object, the robot captures a single frame from an RGB-D sensor. The robot then performs a simple geometric segmentation to extract the points belonging to the object as described in Section~\ref{sec:details}. These 3D points make up the training set $S_{t}$ which the robot uses to estimate the initial object GPIS model as described in Section~\ref{sec:gpis}.

The GPIS model estimated from single a view point cloud typically fails to generate a convex shape. As such, previous approaches to GPIS-based grasp planning~\cite{c8} often take a significant amount of time to find a suitable grasp. To avoid this issue we perform a grasp optimization where the defined cost function encourages the gripper to simply align the center point between the gripper with the center of the object, while constraining the optimization to search over only one degree of freedom in the gripper's orientation. This restricts the total space of grasps allowable for the optimization, while this may make grasping certain objects with obscure geometries infeasible, it works well for common household objects designed to be lifted from the top or opposing sides.

We define the grasping trajectory optimization problem as:
\begin{align}
  \underset{\tau=[\theta_1, \ldots,\theta_H]}{\text{min}}\;
  &p_g\norm{\frac{1}{2}\sum_{q=1}^{2}\phi_{g\_q}(\theta_H) - x_{c}^*}_2^2+\nonumber\\
   &\sum_{k=1}^{H-1}\norm{\frac{1}{2}\sum_{q=1}^{2}\phi_{g\_q}(\theta_k) - x_{k}^f}_2^2 \label{eq:grasp_cost}\\
  \text{s.t.}& \nonumber\\
 &\phi_{g\_q}(\theta_H).\texttt{ang}[i] = x_c^* \; \forall i \neq f \label{eq:orientation-constraint}\\
 &\theta_{min}\preceq\theta_i\preceq\theta_{max} \; \forall i = 1, \ldots, H \label{eq:grasp-joint-limits}\\
 & C(\theta_i,S_{e})\geq 0 \; \forall i = 1, \ldots, H \label{eq:grasp-collision-avoidance}
\end{align}
where $x_{c}^*\in\mathbb{R}^6$ defines the 6D pose corresponding to the centroid of the GPIS model estimated using principle component analysis (PCA)~\cite{c41}. We define the forward kinematics function of the robot arm and gripper as \(\phi_{g\_link}(\theta)\), where $\phi_{g\_1}(\theta)$ and $\phi_{g\_2}(\theta)$ define the forward kinematics for the two gripper jaws respectively. As such the first term in the cost function aligns the gripper with the object at the final timestep, while the second term encourages the end-effector to track a series of task-space waypoints to the grasp.

We define the waypoints for the end-effector to follow in Eq.~(\ref{eq:grasp_cost}) by linearly interpolating the gripper trajectory between the nominal goal position and grasp position $x_f^*$ and initial end-effector pose  $x_0$ to generate a sequence of $H$ waypoints, $x_k^f \in X_f$:
\begin{align}
  x_k^f = \frac{x_f^*-x_0}{H}\cdot k+x_0
  \label{eq:grasp-init}
\end{align}

The constraint defined in Eq.~(\ref{eq:orientation-constraint}) ensures that the final grasp orientation aligns with two of the three directions of the object bounding box; thus only the Euler angle associated with index \(f \in \{1,2,3\}\) change during the optimization. We additionally ignore angle \(f\) in computing the waypoint cost.
\begin{algorithm}[th!]
	\caption{3D GPIS Grasp Planner}
	\label{alg:grasp-planning}
	\begin{algorithmic}[1]
		\REQUIRE ~~\\
		Point cloud $X$, with labels for object points $S_t$\\
		Robot's initial joint state $\theta_0$\\
		\ENSURE ~~\\
		Grasp joint trajectory $\tau^*_g=[\theta_1, \ldots,\theta_H]$
		\STATE Generate object's GPIS model $S_{e}$ from training points $S_{t}$
		\STATE Obtain $x_{c}^* \leftarrow \texttt{PCA}(S_{e})$
		\label{code:fram:trainbase}
		\FOR{$i \leftarrow 1$; $i \leq 3$; $i \leftarrow i+1$ }
                \STATE Compute grasp waypoints $X_f$ using Eq.~(\ref{eq:grasp-init})
		\STATE Compute cost $c_i$ and trajectory $\tau_i$ by solving Eq.~(\ref{eq:grasp_cost}) with orientation dimension \(f=i\)
		\ENDFOR
		\STATE $\tau^* \leftarrow \argmin\{(\tau_1, c_1), (\tau_2, c_2), (\tau_3, c_3)\}$
		\STATE Return $\tau^*$
		\label{code:fram:add}
              \end{algorithmic}
\end{algorithm}

We additionally constrain the optimization to respect the robot joint limits, Eq.~(\ref{eq:grasp-joint-limits}), and avoid collisions with other parts of the object and environment, Eq.~(\ref{eq:grasp-collision-avoidance}). This creates a nonlinear, non-convex optimization. As such we use Sequential Quadratic Programming (SQP) with an inverse-kinematics informed initialization to find the optimal solution.

The signed distance function (SDF) \(\psi(\theta,X)\) computes the signed distance between each mesh of the robot for joint angle \(\theta \in \mathbb{R}^m\) and the given point cloud $X$. We then define the collision constraint as:
\begin{align}
C(\theta,X) = \max(\epsilon-\psi(\theta,X),0)
\end{align}
Which requires the non-gripper links of the robot to maintain some distance \(\epsilon\) from all points in the point cloud.

We define our complete GPIS grasp trajectory optimization in Algorithm~\ref{alg:grasp-planning}. We perform the above optimization three times, allowing each end-effector orientation angle to be free in turn and select the trajectory with lowest cost.

\vspace{-10pt}
\subsection{Reconstruction-Aware Motion Planning}
Our reconstruction-aware trajectory optimization algorithm uses a cost function defined by the conditional entropy of the GPIS model to plan object poses which encourage exploring unseen parts of the object while moving towards the desired object pose. We constrain the optimization to keep the object within a region viewable by the RGB-D sensor. We found that solving this optimization with SQP, even when well initialized, tended to get stuck at local minima, which we attribute to the complexity of the conditional entropy function defined over the GPIS model.

\begin{figure}[thpb]
	\centering
	\includegraphics[scale=0.65]{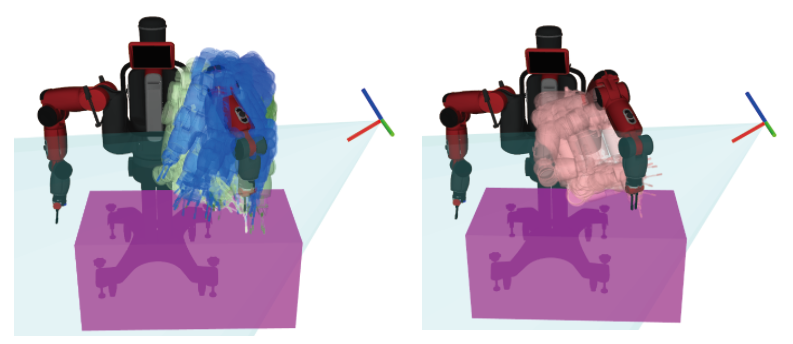}
	\caption{Left: Example training trajectories generated from the GMM. Right: Final trajectory generated through GMM planning. The cyan cone represents the region viewable from the camera; the purple box represents the table obstacle.}
	\label{fig:gmm-train-test}
\end{figure}

As such, we propose using SQP to generate a set of trajectories within the camera range as shown in Fig.~\ref{fig:gmm-train-test} only considering the goal of the transition trajectory as the cost. We then search for an optimal trajectory with respect to the conditional entropy metric through importance sampling, leveraging the approach of Kobilarov~\cite{c42}. We represent the importance density using a Gaussian mixture model (GMM) in a slight variation of~\cite{c42}. This enables us to efficiently solve the trajectory optimization problem, while still generating high-quality trajectories. We assume the goal pose, $x_d\in\mathbb{R}^6$, is within the robot's reachable workspace and the viewpoint of the camera.

Algorithm~\ref{alg:obj-transition} outlines our reconstruction-aware trajectory optimization approach.  Our GMM-based trajectory generation method iteratively approximates the optimal path in the robot configuration space to quickly explore unknown parts of the object. Lines~\ref{code:init_traj_start}--\ref{code:init_traj_end} define the generation of our initial trajectory set. We randomize the object goal’s orientation generating \(N_o\) surrogate goals $x_d^i$, in order to generate a sufficiently diverse initial set of trajectories.
The planner then generates collision-free configuration-space trajectories to reach each goal pose using an SQP approach similar to our grasp planner:
\begin{align}
\underset{\theta=[\theta_1, \ldots,\theta_m]}{\text{min}}
&p_t\norm{\frac{1}{2}\sum_{q=1}^{2}\phi_{g\_q}(\theta_m) - x_d}_2^2+\nonumber\\
&\sum_{j=1}^{m-1}\norm{\frac{1}{2}\sum_{q=1}^{2}\phi_{g\_q}(\theta_j) - x_{j}}_2^2 \label{eq:transition_cost}\\
\text{s.t.}& \nonumber\\
&\theta_{min}\preceq\theta_i\preceq\theta_{max} \;, \forall i = 1, \ldots, M \label{eq:joint-constraints}\\
& C_{cone}(\theta_i)\leq 0 \;, \forall i = 1, \ldots, M \label{eq:cone-constraint} \\
& C(\theta_i)\geq 0 \;, \forall i = 1, \ldots, M \label{eq:opt3}
\end{align}
where again we've used linear interpolation between the initial pose and desired pose to guide the optimization. We add an additional constrains in Eq.~(\ref{eq:cone-constraint}) to ensure the object remains within the region viewable by the camera. We give more details about this constraint in Section~\ref{sec:camera-view}.

Lines~\ref{code:ce-start}--\ref{code:ce-end} perform the cross-entropy motion planning to improve the trajectories based on the expected reconstruction. This iteratively operates through two alternating stages of generating an elite set of trajectories sampled from a GMM and then fitting a new GMM model to this elite set using the expectation-maximization (EM) algorithm. This continues until the converge criteria (Line~\ref{code:ce-start}) are met.  Lines~\ref{code:elite-set-start}--\ref{code:elite-set-end} generate an elite set, \(\eta_j\), based on the conditional entropy of the GPIS model, defined over the entire trajectory \(\tau\). We define our cost as $J(S_{\tau})=-H(S_{c}|S_{t})$. Here $S_{c}$ is the set of GPIS points expected to be captured by camera as described in Algorithm~\ref{alg:raycasting}. This cost guides the planner to select trajectories which allow the robot to explore previously unobserved parts of the object while its moving towards the goal.
\begin{algorithm}[ht]
  \caption{Reconstruction-Aware Trajectory Optimization}
  \label{alg:obj-transition}
  \begin{algorithmic}[1]
    \REQUIRE ~~\\
    GPIS model pose $x_{c}^*$\\
    Object goal pose $x_{d}$\\
    Number of initial random orientations $N_{o}$\\
    Number of GMM components $N_{c}$\\
    Initial joint state $\theta_0$\\
    \ENSURE ~~\\
    Reconstruction-aware joint trajectory $\tau^{*}_{r}=[\theta_1, \ldots,\theta_M]$;\\
    \STATE \(T \leftarrow \{\}\)
    \FOR {\(i \leftarrow 0; i < N_o; i \leftarrow i + 1\)}\label{code:init_traj_start}
    \STATE \(x_d^i \leftarrow x_d\) with random orientation
    \STATE Generate \(\tau_i\) by solving optimization in Eq.~(\ref{eq:transition_cost}) with \(x_d^i\)
    \STATE \(T \leftarrow T \cup \tau_i\)
    \ENDFOR\label{code:init_traj_end}
    \STATE $j \leftarrow 0$; $\gamma_0=\infty$
    \WHILE {Not converged}\label{code:ce-start}
    \STATE \(\eta_j \leftarrow \{\}\)\label{code:elite-set-start}
    \FOR {\(\tau_i \in T\)}
    \STATE \(S_{\tau_{i}} \leftarrow \) Algorithm~\ref{alg:raycasting} \((\tau_i)\)
    \IF {\(J(S_{\tau_{i}}) \leq \gamma_j\)}
    \STATE $\eta_j \leftarrow \eta_j \cup \{\tau_i\}$
    \ENDIF
    \ENDFOR\label{code:elite-set-end}
    \STATE $G_j \leftarrow \texttt{EM}(\eta_j, N_c)$ // Fit GMM using EM \label{code:em}
    \STATE \(T \leftarrow\) Algorithm~\ref{alg:gmm} \((G_j, N_o)\)
    \STATE $\gamma_{j+1} \leftarrow J_q$ the $1^{st}$ quantile score over all \(\tau \in T\)
    \STATE $j \leftarrow j+1$
    \ENDWHILE\label{code:ce-end}
    \STATE \(\tau_r^{*} \leftarrow \argmin{ J(\tau)\; \forall \tau \in T}\)
    \RETURN \(\tau_r^{*}\)
  \end{algorithmic}
\end{algorithm}

Following e.g.~\cite{c22}, the conditional entropy $H(S_{c}|S_{t})$ for the GPIS model equals
\begin{equation}
\begin{aligned}
J(S_{\tau}) = -H(S_{c}|S_{t}) = -\frac{1}{2}\ln((2\pi e)^3 |\Sigma_{S_{c}|S_{t}}|)
\label{entropy}
\end{aligned}
\end{equation}
where $\Sigma_{S_{c}|S_{t}}$ is equal to,
\begin{equation}
\begin{aligned}
\Sigma_{S_{c}|S_{t}}=&K(S_{c},S_{c})+\sigma_nI-  \\
&K(S_{c},S_{t})(K(S_{t},S_{t})+\sigma_nI)^{-1}K(S_{t},S_{c})
\end{aligned}
\end{equation}
where $S_{c} \subseteq S_{e}$ defines the subset of the estimated GPIS points, which are observable by the camera along the trajectory. We can compute $S_{c}$ by ray-casting on the GPIS model as shown in Algorithm~\ref{alg:raycasting}.

Given this elite set, \(\eta_j\), Algorithm~\ref{alg:obj-transition} fits a new GMM using EM (Line~\ref{code:em}) and passes the resulting GMM \(G_j\) to Algorithm~\ref{alg:gmm}.  The view-constrained parameter sampler defined in Alg.~\ref{alg:gmm} uses rejection sampling to generate samples from the GMM \(G\) which obey the trajectory constraints defined in Eq.~\ref{eq:joint-constraints}--\ref{eq:opt3}. Given the new set of trajectories \(T\) Alg.~\ref{alg:obj-transition} then updates the threshold, \(\gamma_j\) for inclusion in the elite set to equal the first quantile score for the generated trajectories.

The algorithm converges when the change in the best-cost trajectory moves below a small threshold, \(\epsilon_c\), between two iterations or the worst cost trajectory in the elite set receives cost better than some cost threshold, \(\delta\).
\begin{algorithm}[htb]
	\caption{View-constrained parameter sampling}
	\label{alg:gmm}
	\begin{algorithmic}[1]
		\REQUIRE ~~\\
		GMM $G = \{(\mu_k, \Sigma_k, \omega_k)\}_{k=1:N_c}$\\
                Number of desired trajectories \(N\)
		\ENSURE ~~\\
		Set of \(N\) joint trajectories \(T = \{\tau_i, \ldots, \tau_n\}\)
                \WHILE {\(i < N\)}
		\STATE Choose $k \in \{1,...,N_c\}$ proportional to $\omega_k$
		\STATE Sample $r \sim \mathcal{N}(0,1)$ and set $Z_i \leftarrow \mu_k + r \cdot \sqrt{\Sigma_k}$
		\IF {$Z_i \in Z_{con}$}
                \STATE \(T \leftarrow T \cup \{Z_i\}\)
                \STATE $i \leftarrow i+1$
                \ENDIF
                \ENDWHILE
                \RETURN \(T\)
	\end{algorithmic}
\end{algorithm}
\vspace{-20pt}
\subsection{Camera View Model}\label{sec:camera-view}
If at any timestep along the trajectory no estimated point cloud can be captured, then we can not compute $|\Sigma_{X_k^*|X}|$ in Eq.~(\ref{entropy}). To avoid this problem we constrain the object to remain within a region viewable by the camera; we approximate this model with a truncated cone as shown in Fig.~\ref{fig:gmm-train-test}.
We then add a constraint on the optimization that the estimated object centroid $x_{c}^*$, remains in the cone. By assuming the object is rigidly attached to the gripper, we can express this constraint using a signed-distance function as \( C_{cone}(\theta_i):\mathbb{R}^m \xrightarrow{}\mathbb{R}\).
In order to speed up the collision checking between the robot meshes and table point cloud, we use a bounding box model to approximate the table in front of the robot, displayed as the semi-transparent purple box in Fig.~\ref{fig:gmm-train-test}.

\begin{algorithm}[htb]
  \caption{Raycasting on GPIS}
  \label{alg:raycasting}
  \begin{algorithmic}[1]
    \REQUIRE ~~\\
    Estimated point cloud set $S_{e}$\\
    Joint trajectory $\tau_{s}=[\theta_1, \ldots,\theta_M]$
    \ENSURE ~~\\
    Set of GPIS points observed during the trajectory $S_{c}$
    \FOR {$\theta_i$ in $\tau_{s}$}
    \STATE \(S_i \leftarrow \) estimated GPIS point cloud for waypoint $\theta_i$
    \FOR {Each ray, \(b_i\) in camera beam model}
    \STATE Raycast on $S_i$ generating observable point $x^*_i$
    \IF {$x^*_i \in S_i$ AND $\mu(x^*_i)\le\eta$}
    \STATE $x^*_q \leftarrow x \in S_{e}$ s.t. \(x^*_i = \phi(\theta_i)\cdot x\)
    \STATE $S_{c} \leftarrow S_{c}\cup\{x^*_q\}$
    \ENDIF
    \ENDFOR
    \ENDFOR
  \end{algorithmic}
\end{algorithm}
%%%%%%%%%%%%%%%%%%%%%%%%%%%%%%%%%%%%%%%%%%%%%%%%%%%%%%%%%%%%%%%%%%%%%%%%%%%%%%%% 
\vspace{-10pt}
\section{Physical-Robot Experiments}\label{sec:experiments}
\begin{figure*}[thpb]
	\centering
	\includegraphics[scale=0.476]{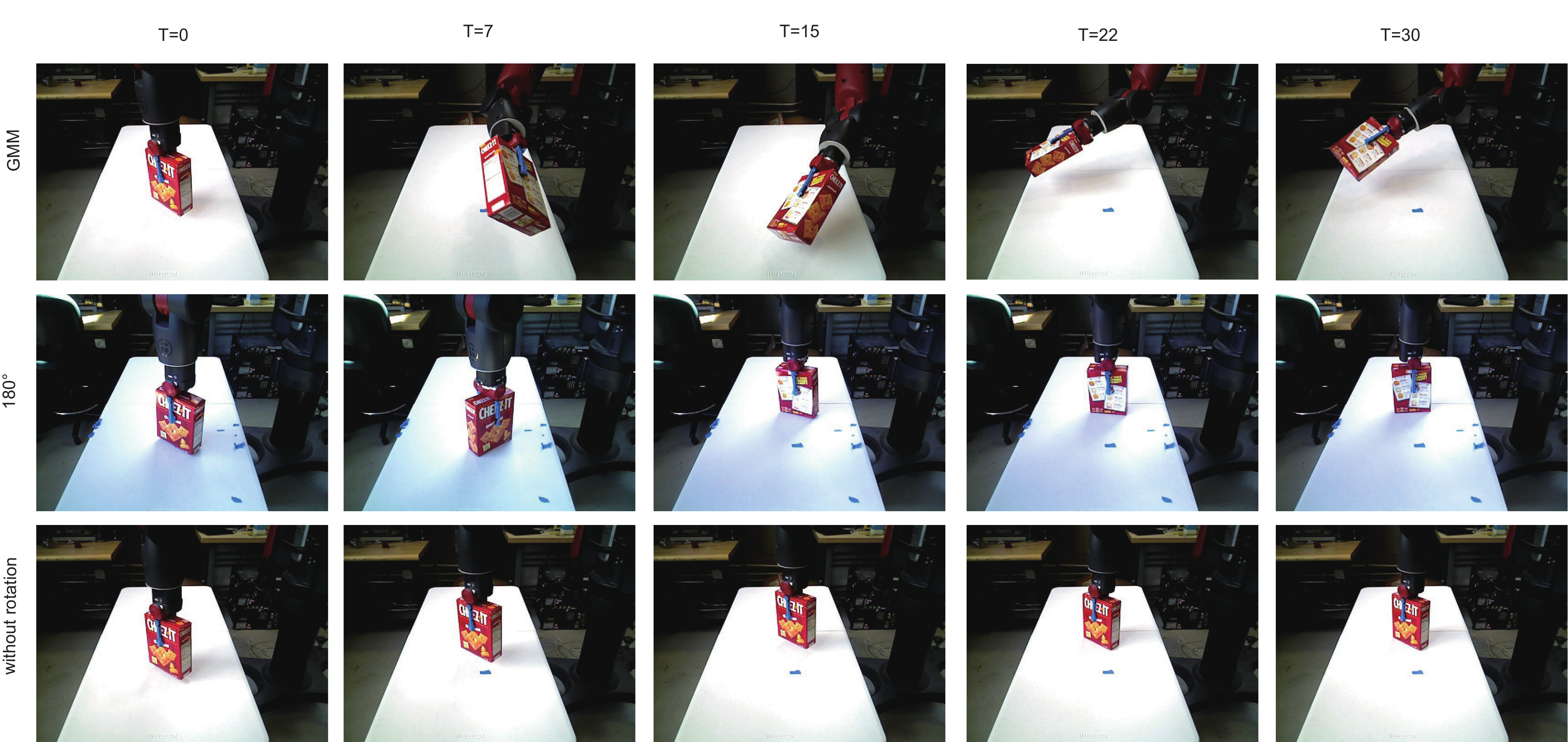}
	\caption{Comparison of trajectories generated by our reconstruction-aware (GMM) planner, a heuristic approach (180\textdegree), and a goal-based trajectory optimization (without rotation). All sides of the cracker box are explored by the reconstruction-aware trajectory while the bottom side is not seen in the alternative approaches.}
	\label{fig:label-trajectory-compare}
\end{figure*}
In this section, we present the implementation details, protocol, and results for experiments performed on our physical-robot platform. We compare the object model reconstructions generated from our reconstruction-aware trajectory optimization and two alternative planning approaches. The software used to perform all experiments is available at \url{https://bitbucket.org/robot-learning/reconstruction-aware_trajectory_optimization/}.

\subsection{Experimental Setup and Implementation Details}\label{sec:details}
Since both the grasping cost function and object transition cost functions are non-convex functions, we use sequential quadratic programming (SQP) to optimize the trajectory. SQP has already shown convincing results in~\cite{sundaralingam-auro2018-in-grasp-optimization} for generating trajectories with sparsely constrained cost functions. We use SNOPT~\cite{c12} to solve all SQP problems.
We implement our GPIS model using GPflow~\cite{c19}, a TensorFlow-based Gaussian process library to learn the initial GPIS model, estimate whether a point lies on the surface, and compute the covriance of estimated points. For the purpose of saving limited GPU memory, we augment this with GPy, a Gaussian process framework in python, to compute the derivative of the covariance with respect to the estimated points during planning.

\begin{figure*}[thpb]
	\centering
	\includegraphics[scale=0.7]{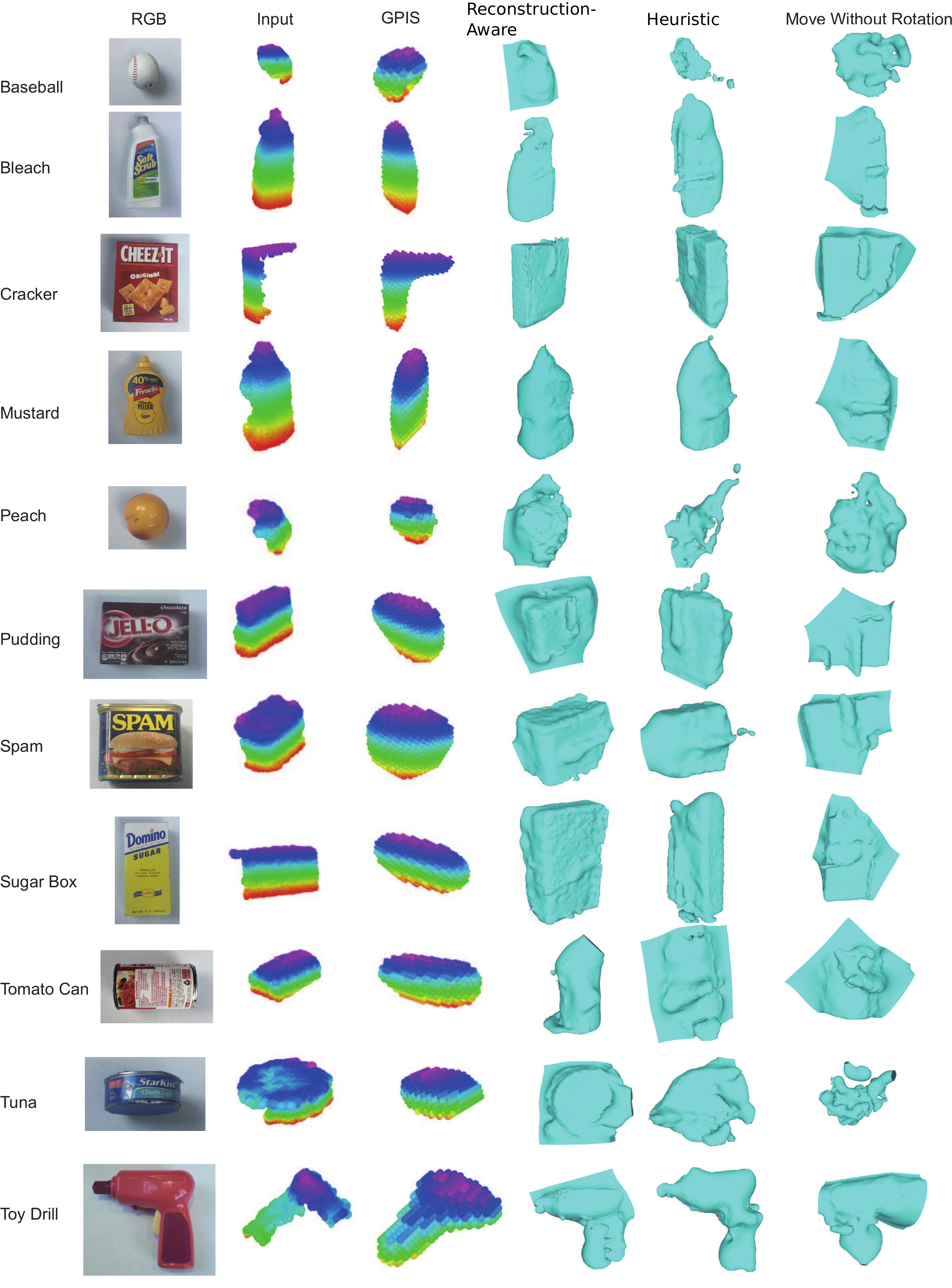}
	\caption{Comparison of object reconstruction results for all 11 objects. The first column shows the objects. The second column visualizes the initial object point cloud segmentation; while the third  visualizes the initial GPIS model. The last three column show the reconstruction results for the trajectory-aware, heuristic, and goal-directed planning approaches.}
	\label{fig:reconstruction-viz}
\end{figure*}

We use Baxter research robot, equipped with two 7-DOF arms and two parallel-jaw grippers with ROS Indigo~\cite{c29} for all experiments.
The gripper width of the Baxter is approximately 460 mm, and we pick ten objects suitable for the gripper from YCB dataset, which are baseball, bleach, cracker box, mustard, peach, pudding, spam, sugar box, tomato can, and toy drill. All selected objects are non-transparent for compatibility with the RGB-D sensor. The first column of Fig.~\ref{fig:reconstruction-viz} shows images of all objects.
\begin{figure}[thpb]
	\centering
	\includegraphics[width=\linewidth]{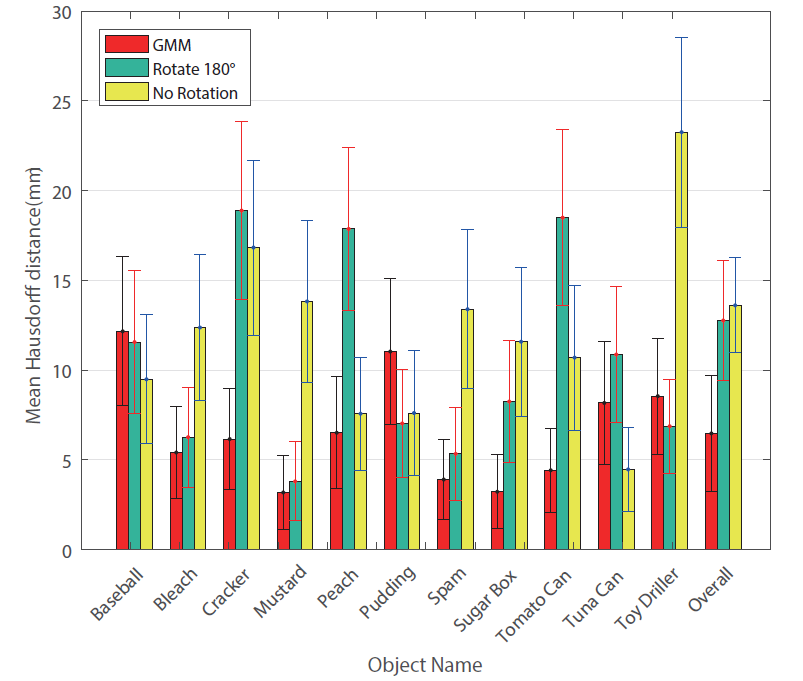}
	\caption{Comparison of reconstruction error for the different planning approaches. GMM denotes the reconstruction-aware trajectory. Error bars report standard deviation.}
	\label{fig:reconstruction-error}
\end{figure}

To evaluate the grasping performance using our GPIS representation, we performed a number of test grasps evaluating the hyper-parameters. We empirically determined the values of $p_g=0.5$ in Eq.~(\ref{eq:grasp_cost}) and $\eta = 0.1$ in Eq.~(\ref{eq:implicit-surf}). We found with these values the resulting GPIS model tended to be closed.
We empirically determined the value of $p_t=0.5$ in Eq.~(\ref{eq:transition_cost}). In running the trajectory-aware reconstruction the number of training trajectories for the GMM is $N_O=50$ and the number of components is two.

We use Co-Fusion~\cite{c39} to reconstruct the object model offline. We use the DART tracker~\cite{c30} to track and remove the robot arm from the RGB-D point cloud. We additionally remove the points associated with the table by fitting a plane using RANSAC~\cite{fischler-commacm1981}. After running the trajectory and fusing the point cloud, we use Poisson Surface Reconstruction~\cite{Kazhdan-SGP06} to generate the final object mesh.

\subsection{Object Manipulation Experiments}
We evaluate the object reconstruction performance of our reconstruction-aware trajectory optimization by comparing to two alternative trajectory approaches. The first alternative implements a heuristic approach that simply rotates Baxter's final joint {180\textdegree} before moving the object to the goal. The second alternative directly plans a trajectory to move the object from the initial grasp to the goal location not taking reconstruction into account. Figure~\ref{fig:label-trajectory-compare} shows example trajectories for the three different planning approaches. As we can see, the cracker box is fully explored through the reconstruction-aware trajectory, while the bottom side of the box is never captured by the RGB-D sensor in the other approaches, even though the heuristic approach does capture the backside of the box. These trajectories are representative of what was seen for all objects.

In order to qualitative compare the approaches, we compute error using the Hausdorff distance between the surface of the reconstructed mesh to the surface of the ground-truth YCB mesh. We computed the Hausdorff distance using Meshlab’s Hausdorff distance filter in a single direction. Figure~\ref{fig:reconstruction-error} shows the mean values and standard deviation of the Hausdorff distance across all points on the mesh for the reconstructions generated with the three different methods for trajectory optimization. From the results, we can infer that our reconstruction-aware trajectories perform better than rotating Baxter's final joint {180\textdegree} except on the pudding box, baseball, and tuna can. The reason that those objects' reconstruction error are larger than the other two methods is they are small objects and the RGB-D sensor's resolution is not high enough to generate clear point clouds. This results in difficulty for the fusion algorithm to reconstruct the object. For the toy drill, due to the lack of richness of color features, the fusion algorithm does not reconstruct the object mesh corresponding to the reconstruction-aware trajectory better than the one to rotating {180\textdegree}. This could potentially be improved using richer appearance-based cues for tracking the object.

Finally we show the reconstructed surfaces for all objects using all three approaches in Fig.~\ref{fig:reconstruction-viz}. While not perfect, we note that reconstructions generated from the reconstruction-aware trajectories show more object detail and better match the global object shape than the alternative approaches.

%%%%%%%%%%%%%%%%%%%%%%%%%%%%%%%%%%%%%%%%%%%%%%%%%%%%%%%%%%%%%%%%%%%%%%%%%%%%%%%%
\section{Conclusions and Future Work}\label{sec:conclusion}
We presented a novel framework for jointly grasping and reconstructing an unknown objectusing a Gaussian process implicit surface model. We combine nonlinear optimization and sampling based motion planning algorithms to explore the unknown object during a pick-and-place trajectory and reconstruct the final 3D model using a structure from motion algorithm. Experiments show that pick-and-place trajectories generated using our approach generate higher quality object models than simply rotating the final joint for {180\textdegree} or planning a motion that does not account for viewpoint.

Many interesting directions exist for future work. First we wish to examine how a single optimization can plan both the grasp and object motion to better account for the occlusion of the gripper pose and dealing with additional motion constraints imposed by clutter. Second, we wish to incorporate tactile sensing to improve the 3D reconstruction such as in~\cite{c9,c20} while also estimating material properties of the object~\cite{hoelscher-ichr2015-tactile-recoginition}, and inertial parameters (i.e. mass, center of mass, inertia). Finally, we wish to explore planning to improve object modeling during more complex manipulation tasks such as in-hand regrasping~\cite{sundaralingam-auro2018-in-grasp-optimization,sundaralingam-icra2018-finger-gaiting}.
%%%%%%%%%%%%%%%%%%%%%%%%%%%%%%%%%%%%%%%%%%%%%%%%%%%%%%%%%%%%%%%%%%%%%%%%%%%%%%%%
\vspace{-10pt}
\bibliography{entropy}

\begin{thebibliography}{10}
\providecommand{\url}[1]{#1}
\csname url@rmstyle\endcsname
\providecommand{\newblock}{\relax}
\providecommand{\bibinfo}[2]{#2}
\providecommand\BIBentrySTDinterwordspacing{\spaceskip=0pt\relax}
\providecommand\BIBentryALTinterwordstretchfactor{4}
\providecommand\BIBentryALTinterwordspacing{\spaceskip=\fontdimen2\font plus
\BIBentryALTinterwordstretchfactor\fontdimen3\font minus
  \fontdimen4\font\relax}
\providecommand\BIBforeignlanguage[2]{{%
\expandafter\ifx\csname l@#1\endcsname\relax
\typeout{** WARNING: IEEEtran.bst: No hyphenation pattern has been}%
\typeout{** loaded for the language `#1'. Using the pattern for}%
\typeout{** the default language instead.}%
\else
\language=\csname l@#1\endcsname
\fi
#2}}

\bibitem{sundaralingam-auro2018-in-grasp-optimization}
B.~Sundaralingam and T.~Hermans, ``{Relaxed-Rigidity Constraints: Kinematic
  Trajectory Optimization and Collision Avoidance for In-Grasp Manipulation},''
  \emph{{Autonomous Robots}}, 2018.

\bibitem{c26}
J.~Ilonen, J.~Bohg, and V.~Kyrki, ``Three-dimensional object reconstruction of
  symmetric objects by fusing visual and tactile sensing,'' \emph{Int. Journal
  of Robotics Research}, vol.~33, no.~2, pp. 321--341, 2014.

\bibitem{c27}
M.~Meier, M.~Schopfer, R.~Haschke, and H.~Ritter, ``A probabilistic approach to
  tactile shape reconstruction,'' \emph{in IEEE Transactions on Robotics},
  vol.~27, no.~3, pp. 630--635, 2011.

\bibitem{c28}
F.~Endres, J.~Hess, J.~Sturm, D.~Cremers, and W.~Burgard, ``3-d mapping with an
  rgb-d camera,'' \emph{in IEEE Transactions on Robotics}, vol.~30, no.~1, pp.
  177--187, 2014.

\bibitem{c33}
J.~Zhou, R.~Paolini, A.~M. Johnson, J.~A. Bagnell, and M.~T. Mason, ``A
  probabilistic planning framework for planar grasping under uncertainty,''
  \emph{in IEEE Robotics and Automation Letters}, vol.~2, no.~4, pp.
  2111--2118, 2017.

\bibitem{c25}
J.~Varley, D.~Watkins, and P.~Allen, ``Visual-tactile geometric reasoning,'' in
  \emph{Robotics: Science and Systems Workshop}, 2012, pp. 4783--4790.

\bibitem{c11}
B.~Calli, A.~Singh, A.~Walsman, S.~Srinivasa, P.~Abbeel, and A.~M. Dollar,
  ``The ycb object and model set: Towards common benchmarks for manipulation
  research,'' in \emph{IEEE Int. Conf. Advanced Robotics}, 2015, pp. 510--517.

\bibitem{c35}
H.~Mao, Z.~Teng, and J.~Xiao, ``Progressive object modeling with a continuum
  manipulator in unknown environments,'' in \emph{IEEE Int. Conf. Robotics and
  Automation}, 2017, pp. 5674--5681.

\bibitem{c36}
S.~Kriegel, M.~Brucker, Z.-C. Marton, T.~Bodenmuller, and M.~Suppa, ``Combining
  object modeling and recognition for active scene exploration,'' in
  \emph{IEEE/RSJ Int. Conf. on Intelligent Robots and Systems}, 2013, pp.
  2384--2391.

\bibitem{c37}
J.~C. Caicedo and S.~Lazebnik, ``Active object localization with deep
  reinforcement learning,'' in \emph{IEEE Int. Conf. on Computer Vision}, 2015,
  pp. 2488--2496.

\bibitem{c38}
E.~Herbst, P.~Henry, and D.~Fox, ``Toward online 3-d object segmentation and
  mapping,'' in \emph{IEEE Int. Conf. Robotics and Automation}, 2014, pp.
  3193--3200.

\bibitem{c13}
B.~Charrow, S.~Liu, V.~Kumar, and N.~Michael, ``Information-theoretic mapping
  using cauchy-schwarz quadratic mutual information,'' in \emph{IEEE Int. Conf.
  Robotics and Automation}, 2015, pp. 4791--4798.

\bibitem{c14}
B.~J. Julian, S.~Karaman, and D.~Rus, ``On mutual information-based control of
  range sensing robots for mapping applications,'' \emph{Int. Journal of
  Robotics Research}, vol.~33, no.~10, pp. 1375--1392, 2014.

\bibitem{c17}
J.~Xie, Y.-F. Hsu, R.~S. Feris, and M.-T. Sun, ``Fine registration of 3d point
  clouds fusing structural and photometric information using an rgb-d camera,''
  \emph{in Journal of Visual Communication and Image Representation}, vol.~32,
  pp. 194--204, 2015.

\bibitem{c32}
K.~Lai, L.~Bo, X.~Ren, and D.~Fox, ``A large-scale hierarchical multi-view
  rgb-d object dataset,'' in \emph{IEEE Int. Conf. Robotics and Automation},
  2011, pp. 1817--1824.

\bibitem{c18}
R.~A. Newcombe, S.~Izadi, O.~Hilliges, D.~Molyneaux, D.~Kim, A.~J. Davison,
  P.~Kohi, J.~Shotton, S.~Hodges, and A.~Fitzgibbon, ``Kinectfusion: Real-time
  dense surface mapping and tracking,'' in \emph{IEEE Int. Sym. on Mixed and
  Augmented Reality}, 2011, pp. 127--136.

\bibitem{c15}
M.~Krainin, P.~Henry, X.~Ren, and D.~Fox, ``Manipulator and object tracking for
  in-hand 3d object modeling,'' \emph{Int. Journal of Robotics Research},
  vol.~30, no.~11, pp. 1311--1327, 2011.

\bibitem{krainin-icra2011-nbv}
M.~Krainin, B.~Curless, and D.~Fox, ``{Autonomous Generation of Complete 3D
  Object Models Using Next Best View Manipulation Planning},'' in \emph{IEEE
  Int. Conf. Robotics and Automation}, 2011.

\bibitem{c8}
J.~Mahler, S.~Patil, B.~Kehoe, J.~Van Den~Berg, M.~Ciocarlie, P.~Abbeel, and
  K.~Goldberg, ``Gp-gpis-opt: Grasp planning with shape uncertainty using
  gaussian process implicit surfaces and sequential convex programming,'' in
  \emph{IEEE Int. Conf. Robotics and Automation}, 2015, pp. 4919--4926.

\bibitem{c9}
S.~Dragiev, M.~Toussaint, and M.~Gienger, ``Gaussian process implicit surfaces
  for shape estimation and grasping.'' in \emph{IEEE Int. Conf. Robotics and
  Automation}, 2011, pp. 2845--2850.

\bibitem{c20}
Z.~Yi, R.~Calandra, F.~Veiga, H.~van Hoof, T.~Hermans, Y.~Zhang, and J.~Peters,
  ``Active tactile object exploration with gaussian processes,'' in
  \emph{IEEE/RSJ Int. Conf. on Intelligent Robots and Systems}, 2016, pp.
  4925--4930.

\bibitem{c1}
M.~Bjorkman, Y.~Bekiroglu, V.~Hogman, and D.~Kragic, ``Enhancing visual
  perception of shape through tactile glances,'' in \emph{IEEE/RSJ Int. Conf.
  on Intelligent Robots and Systems}, 2013, pp. 3180--3186.

\bibitem{c2}
J.~Ilonen, J.~Bohg, and V.~Kyrki, ``Fusing visual and tactile sensing for 3-d
  object reconstruction while grasping,'' in \emph{IEEE Int. Conf. Robotics and
  Automation}, 2013, pp. 3547--3554.

\bibitem{c4}
D.~Chen, V.~Dietrich, and G.~von Wichert, ``Precision grasping based on
  probabilistic models of unknown objects,'' in \emph{IEEE Int. Conf. Robotics
  and Automation}, 2016, pp. 2044--2051.

\bibitem{bohg-tro2017-interactive-perception}
J.~{Bohg}, K.~{Hausman}, B.~{Sankaran}, O.~{Brock}, D.~{Kragic}, S.~{Schaal},
  and G.~S. {Sukhatme}, ``Interactive perception: Leveraging action in
  perception and perception in action,'' \emph{IEEE Transactions on Robotics},
  vol.~33, no.~6, pp. 1273--1291, Dec 2017.

\bibitem{c23}
T.~Hermans, J.~M. Rehg, and A.~Bobick, ``Guided pushing for object
  singulation,'' in \emph{IEEE/RSJ Int. Conf. on Intelligent Robots and
  Systems}, 2012, pp. 4783--4790.

\bibitem{c34}
H.~Van~Hoof, O.~Kroemer, and J.~Peters, ``Probabilistic segmentation and
  targeted exploration of objects in cluttered environments,'' \emph{in IEEE
  Transactions on Robotics}, vol.~30, no.~5, pp. 1198--1209, 2014.

\bibitem{c5}
L.~Ma, M.~Ghafarianzadeh, D.~Coleman, N.~Correll, and G.~Sibley, ``Simultaneous
  localization, mapping, and manipulation for unsupervised object discovery,''
  in \emph{IEEE Int. Conf. Robotics and Automation}, 2015, pp. 1344--1351.

\bibitem{c16}
W.~Martens, Y.~Poffet, P.~R. Soria, R.~Fitch, and S.~Sukkarieh, ``Geometric
  priors for gaussian process implicit surfaces,'' \emph{in IEEE Robotics and
  Automation Letters}, vol.~2, no.~2, pp. 373--380, 2017.

\bibitem{c6}
Y.~Ren, ``Implicit shape representation for 2d/3d tracking and
  reconstruction,'' Ph.D. dissertation, University of Oxford, 2014.

\bibitem{c21}
M.~Seeger, ``Gaussian processes for machine learning,'' \emph{Int. journal of
  neural systems}, vol.~14, no.~02, pp. 69--106, 2004.

\bibitem{c31}
R.~B. Rusu and S.~Cousins, ``{3D is Here: Point Cloud Library (PCL)},'' in
  \emph{IEEE Int. Conf. Robotics and Automation}, 2011, pp. 1--4.

\bibitem{c41}
S.~Wold, K.~Esbensen, and P.~Geladi, ``Principal component analysis,''
  \emph{Chemometrics and intelligent laboratory systems}, vol.~2, no. 1-3, pp.
  37--52, 1987.

\bibitem{c42}
M.~Kobilarov, ``{Cross-entropy motion planning},'' \emph{Int. Journal of
  Robotics Research}, vol.~31, no.~7, pp. 855--871, 2012.

\bibitem{c22}
J.~C. Principe, \emph{Information theoretic learning: Renyi's entropy and
  kernel perspectives}.\hskip 1em plus 0.5em minus 0.4em\relax Springer Science
  \& Business Media, 2010.

\bibitem{c12}
P.~E. Gill, W.~Murray, and M.~A. Saunders, ``Snopt: An sqp algorithm for
  large-scale constrained optimization,'' \emph{SIAM review}, vol.~47, no.~1,
  pp. 99--131, 2005.

\bibitem{c19}
D.~G. Matthews, G.~Alexander, M.~Van Der~Wilk, T.~Nickson, K.~Fujii,
  A.~Boukouvalas, P.~Le{\'o}n-Villagr{\'a}, Z.~Ghahramani, and J.~Hensman,
  ``Gpflow: A gaussian process library using tensorflow,'' \emph{Journal of
  Machine Learning Research}, vol.~18, no.~1, pp. 1299--1304, 2017.

\bibitem{c29}
M.~Quigley, K.~Conley, B.~Gerkey, J.~Faust, T.~Foote, J.~Leibs, R.~Wheeler, and
  A.~Y. Ng, ``Ros: an open-source robot operating system,'' in \emph{ICRA
  workshop on open source software}, vol.~3, no. 3.2, 2009, p.~5.

\bibitem{c39}
M.~R{\"u}nz and L.~Agapito, ``Co-fusion: Real-time segmentation, tracking and
  fusion of multiple objects,'' in \emph{IEEE Int. Conf. Robotics and
  Automation}, 2017, pp. 4471--4478.

\bibitem{c30}
T.~Schmidt, R.~A. Newcombe, and D.~Fox, ``Dart: Dense articulated real-time
  tracking.'' in \emph{Robotics: Science and Systems}, vol.~2, no.~1, 2014.

\bibitem{fischler-commacm1981}
M.~A. Fischler and R.~C. Bolles, ``Random sample consensus: a paradigm for
  model fitting with applications to image analysis and automated
  cartography,'' \emph{Commun. ACM}, vol.~24, June 1981.

\bibitem{Kazhdan-SGP06}
\BIBentryALTinterwordspacing
M.~Kazhdan, M.~Bolitho, and H.~Hoppe, ``{Poisson Surface Reconstruction},'' in
  \emph{Eurographics Symposium on Geometry Processing}.\hskip 1em plus 0.5em
  minus 0.4em\relax Aire-la-Ville, Switzerland, Switzerland: Eurographics
  Association, 2006, pp. 61--70. [Online]. Available:
  \url{http://dl.acm.org/citation.cfm?id=1281957.1281965}
\BIBentrySTDinterwordspacing

\bibitem{hoelscher-ichr2015-tactile-recoginition}
J.~Hoelscher, J.~Peters, and T.~Hermans, ``{Evaluation of Tactile Feature
  Extraction for Interactive Object Recognition},'' in \emph{{IEEE-RAS Int.
  Conference on Humanoid Robotics}}, 2015.

\bibitem{sundaralingam-icra2018-finger-gaiting}
B.~Sundaralingam and T.~Hermans, ``{Geometric In-Hand Regrasp Planning:
  Alternating Optimization of Finger Gaits and In-Grasp Manipulation},'' in
  \emph{{IEEE Int. Conf. on Robotics and Automation}}, 2018.

\end{thebibliography}
\bibliographystyle{IEEEtran}

\end{document}